\documentclass[graybox]{svmult}

\usepackage{mathptmx}       
\usepackage{helvet}         
\usepackage{courier}        
\usepackage{type1cm}        
\usepackage{makeidx}         
\usepackage{graphicx}        
\usepackage{multicol}        
\usepackage[bottom]{footmisc}

\usepackage{natbib}
\usepackage{multirow}
\usepackage{hyperref}

\makeindex             

\begin{document}

\title*{Lexicase Selection Parameter Analysis: Varying Population Size and Test Case Redundancy with Diagnostic Metrics} 
\titlerunning{Lexicase Selection with Varying Population Size and Test Case Uniformity}
\author{Jose Guadalupe Hernandez, Anil Kumar Saini, Jason H. Moore}
\authorrunning{Hernandez, Saini, Moore}
\institute{
Jose Guadalupe Hernandez \at Cedars-Sinai Medical Center, Los Angeles, CA, USA \email{jose.hernandez8@cshs.org} \and 
Anil Kumar Saini \at Cedars-Sinai Medical Center, Los Angeles, CA, USA \email{anil.saini@cshs.org} \and
Jason H. Moore \at Cedars-Sinai Medical Center, Los Angeles, CA, USA \email{jason.moore@csmc.edu}
}
%
%
\maketitle

\abstract*{Each chapter should be preceded by an abstract (10--15 lines long) that summarizes the content. The abstract will appear \textit{online} at \url{www.SpringerLink.com} and be available with unrestricted access. This allows unregistered users to read the abstract as a teaser for the complete chapter. As a general rule the abstracts will not appear in the printed version of your book unless it is the style of your particular book or that of the series to which your book belongs.
Please use the 'starred' version of the new Springer \texttt{abstract} command for typesetting the text of the online abstracts (cf. source file of this chapter template \texttt{abstract}) and include them with the source files of your manuscript. Use the plain \texttt{abstract} command if the abstract is also to appear in the printed version of the book.}

\abstract{
Lexicase selection is a successful parent selection method in genetic programming that has outperformed other methods across multiple benchmark suites. 
Unlike other selection methods that require explicit parameters to function, such as tournament size in tournament selection, lexicase selection does not.
However, if evolutionary parameters like population size and number of generations affect the effectiveness of a selection method, then lexicase's performance may also be impacted by these `hidden' parameters.
Here, we study how these hidden parameters affect lexicase's ability to exploit gradients and maintain specialists using diagnostic metrics.
By varying the population size with a fixed evaluation budget, we show that smaller populations tend to have greater exploitation capabilities, whereas larger populations tend to maintain more specialists. 
We also consider the effect redundant test cases have on specialist maintenance, and find that high redundancy may hinder the ability to optimize and maintain specialists, even for larger populations.
Ultimately, we highlight that population size, evaluation budget, and test cases must be carefully considered for the characteristics of the problem being solved.
}

\section{Introduction}
\label{sec:1}

Genetic programming has successfully solved problems from various domains such as program synthesis, automated machine learning, and multi-objective optimization.
However, multiple components, procedures, and parameters must be tailored for a given problem to increase the likelihood of a successful genetic programming run.
For example, decisions on solution representation, variation operators, selection methods, population size, and number of generations are typically required for genetic programming.
Additionally, each component or procedure may contain its own set of parameters to consider, such as determining the type of crossover (\textit{e.g.}, one-point, alternation, \textit{etc.}).
Ultimately, all the decisions made in designing a genetic programming system compound and influence the likelihood of success.



A variety of parent selection methods (selection schemes) exists within genetic programming to identify promising solutions to continue evolving.
Some of these methods require an explicit set of parameters to operate.
Tournament selection, for instance, requires a tournament size to identify parents, whereas other methods like fitness-proportionate selection and lexicase selection are \textit{parameter-free}. 
In practice, selection methods are typically implemented as functions in programming parlance that have access to a population of evaluated individuals.
Given the programming framing, all selection methods at a minimum require these two `hidden' parameters: (1) population and (2) solution performances.
Both of these parameters influence a selection method: population size determines the number of potential options and performances determine the priority (quality) of a solution.

Redundancy in test cases can also affect the behavior of a parent selection method. 
For example, methods that aggregate performances will be biased toward redundant test cases, as they are overrepresented in the total sum.
This bias can also occur within methods that do not directly aggregate performances, such as lexicase selection.
In fact, duplicate test cases are more likely to appear early when the set of test cases is randomly shuffled.
As such, lexicase will disproportionately prefer individuals who are best on the test cases with multiple duplicates.
In such a case, more selection events (bigger population size, more generations, or both) are needed to offset this effect.
In real-world problems, redundancy can present itself as duplicate cases in the test set, and as similar test cases such that the performances of solutions on these test cases would be highly correlated.
In this work, we study the dynamics of lexicase with the first type of redundancy and provide insights for future research directions regarding the second type of redundancy.



The ability to intuitively understand how a parent selection method guides an evolutionary search is critical to understanding the problems a particular method is best suited for.
As such, we use the DOSSIER suite from \cite{hernandez2023suite} to understand how population size affects lexicase selection's ability to exploit gradients and maintain specialists given an evaluation budget.
We found that smaller populations facilitate more exploitation, as smaller populations are given more generations to evolve and accumulate mutations.
Conversely, we found that larger population sizes can maintain and optimize more specialists.
However, larger population sizes may negatively impact specialist maintenance when redundant test cases are also used to identify parents.
Ultimately, we illustrate the importance of carefully considering population size, evaluation budget, and test cases when using lexicase selection.

\section{Lexicase Selection Analysis}

\subsection{Lexicase Selection}

Many parent selection methods, such as tournament and fitness proportionate selection, typically use the total error on specific test cases to identify parent solutions.
Lexicase selection~\citep{helmuth2014solving}, however, considers the error on each test case separately. 
The following process is used to select one parent from the entire population. 
First, the set of test cases is randomly shuffled. 
Then, the individuals who perform the best on the first test case are kept in the pool of candidates and the rest are removed. 
The next test case in the sequence is considered and the same process is followed. 
We repeat these steps until only one individual remains in the candidate pool, or no test cases remain. 
In the latter case, we randomly choose an individual from the remaining to become a parent.

Other variants of lexicase have also been proposed. 
For example, epsilon-lexicase selection~\citep{la2016epsilon}  is used for problems where the fitness values are real-valued. 
In down-sampled lexicase and its variants~\citep{hernandez2019random,boldi2024informed}, for every selection event, only a subset of test cases is used to select a parent.
While these variants of lexicase have proven themselves effective, we focus on the original lexicase selection in this work.

\subsection{Analyzing Performance}


Lexicase selection has been used successfully in a variety of evolutionary computation domains. 
Program synthesis is one such domain.
It is the process of generating computer programs that satisfy a given set of constraints or examples, usually given in the form of input/output pairs. 
Lexicase outperformed other parent selection methods (\textit{e.g.}, tournament selection) on problems from from the General Program Synthesis Program Suites (PSB1 and PSB2)~\citep{helmuth_general_2015,helmuth2020benchmarking}. 
In a genetic algorithm setting to solve Boolean Constraint Satisfaction problems, lexicase outperformed fitness-proportionate and tournament selection~\citep{metevier2019lexicase}. 
In Automated Machine Learning (AutoML), lexicase converged to high-performing machine learning pipelines faster than Non-dominated Sorting Genetic Algorithm (NSGA-II)~\citep{matsumoto2023faster}. 


Various works have tried to explain lexicase's improved performance over other parent selection methods. 
Lexicase's ability to select and maintain `specialists' is attributed as a key reason for its performance \citep{helmuth2020importance}.
Specialists are individuals who have low errors on certain test cases with possibly a large overall error.
As such, specialists are less frequently selected by tournament selection and other error-aggregating parent selection methods.
Another explanation for lexicase's performance is that it selects individuals residing on the corners of a high dimensional objective space \citep{la2016epsilon}. 
Using diagnostics problems, \cite{hernandez2023suite,hernandez2022exploration} found that lexicase is better at search space exploration than tournament selection.
In terms of error diversity (number of unique error vectors in the population), lexicase seems to maintain a high diversity compared to tournament selection in program synthesis problems~\citep{helmuth2016effects}. 


Although lexicase selection does not have any parameters associated with the selection step, its performance depends on the underlying evolutionary parameter settings, such as population size and number of generations. 
In effect, we claim that these are some ``hidden'' parameters associated with lexicase selection.
In \cite{la2019probabilistic}, the authors study the effect of population size and number of test cases on the performance of lexicase selection for symbolic regression problems. 
Although the test set size affects the performance of lexicase, the effect is minimal for many problems, at least for the case when lexicase is applied in grammar-guided genetic programming setting~\citep{schweim2022effects}.
In \cite{shahbandegan2024robustness}, the authors find a combination of population size and problem dimension where lexicase fails to optimize contradictory objectives.

\section{Selection scheme diagnostics}
\label{sec:diagnostics}


We use the DOSSIER suite from \cite{hernandez2023suite} to analyze how population size affects lexicase selection under different search spaces and problem conditions.
The suite provides a set of diagnostic tools to characterize a selection scheme's exploitation and exploration abilities.
Here, selection schemes and parent selection methods are synonymous.
The DOSSIER suite comprises eight diagnostics, each using a unique search space with specific problem-solving characteristics (exploitation, exploration, and fitness-valley crossing).
This suite has been used to illustrate differences between several lexicase variants \citep{hernandez2022exploration,shahbandegan2024robustness}, understand the theoretical limits of lexicase \citep{shahbandegan2023theoretical}, and phylogenetic work within evolutionary computation \citep{hernandez2022phylogenetic,lalejini2024phylogeny,shahbandegan2022untangling}.


This work uses the exploitation rate diagnostic and the contradictory objectives diagnostic.
Each diagnostic specifies a unique genotype to phenotype transformation; both the genotype and phenotype are numerical vectors of the same dimensionality.
We refer to each value in a genotype as a gene and each value in a phenotype as a trait.
Solutions consist of genotypes and are assigned phenotypes post-diagnostic evaluation. 
Selection schemes use the phenotype to identify parents (\textit{e.g.}, tournament selection can use an aggregated phenotype).
The evolutionary framework outlined in \cite{hernandez2023suite} isolates lexicase as the primary mechanism for success on a given diagnostic by limiting search space traversal to hill-climbing via only mutations.
This restriction pressures a selection scheme to identify the best set of parents that will lead to success on a given diagnostic.

\subsection{Exploitation rate diagnostic}
\label{sec:exploitation-rate-diagnostic}


The exploitation rate diagnostic measures how fast a selection scheme can steer a population toward a single optimum.
The optimum here refers to a phenotype comprised of traits with the largest numerical value possible.
Efficiently reaching an optimum is particularly crucial for problems with computationally expensive evaluation procedures.
The search space for this diagnostic consists of a smooth, uphill gradient that leads directly toward the single optimum.
To generate a solution's phenotype,  each gene’s value from a solution's genotype is directly copied into the corresponding trait position: the first value of the first gene is copied into the first trait, the value of the second gene is copied into the second trait, and so on.
More specifically, phenotypes are copies of the evaluated genotypes.


This diagnostic favors selection schemes that prioritize overall high-performing solutions, as parents exhibiting this characteristic are positioned closer to the optimum.
Consequently, offspring from these parents can benefit from mutations pushing them toward the optimum.
Truncation selection and tournament selection perform well on this diagnostic as they both aggregate all of a solution's traits and exhibit strong selection pressure for top-performing solutions \citep{hernandez2023suite}.
Lexicase selection, however, has no explicit mechanism for selecting these parents.
Instead, it selects specialists performing well on a subset of test cases, aiming to evolve solutions that eventually perform well on all cases (traits).


When working with a fixed evaluation budget, an evolutionary search must balance the number of generations and the population size.
A larger population requires more parents to be identified per generation, which increases the chances of generating permutations of test cases to identify better-performing solutions.
Moreover, a larger population will result in fewer generations when working with a fixed evaluation budget.
Conversely, a smaller population increases the number of generations, which can lead to the accumulation of more mutations.
Ultimately, this diagnostic allows us to measure how exploitation is affected by these two conflicting dynamics: (1) increasing population size to identify better-performing solutions and (2) increasing the number of generations to accumulate mutations.

\subsection{Contradictory objective diagnostic}
\label{sec:contradictory-objectives-diagnostic}


The contradictory objectives diagnostic captures how well a selection scheme can simultaneously pursue and reach multiple, identical optima within a population.
The ability to reach multiple optima for a given problem is crucial for problems where no single solution exists and problems with conflicting objectives.
The search space for this diagnostic consists of multiple identical, smooth, uphill gradients leading to unique optima.
The diagnostic dimensionality determines the number of gradients and optima in the search space.

\begin{figure}[!h]
\centering
\includegraphics[width=0.9\textwidth]{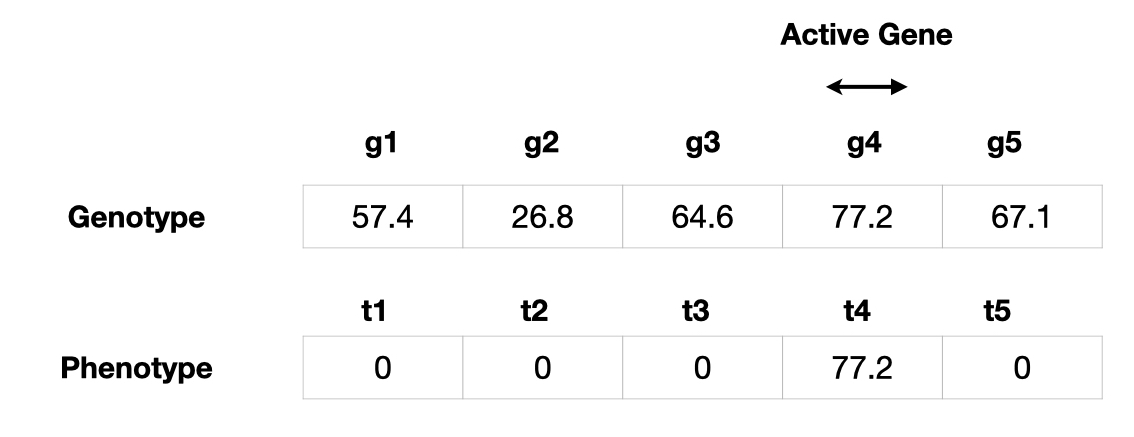}
\caption{Example phenotype construction for the contradictory objectives diagnostic. Note that the trait values of all genes except for the maximum value are zero. In this case, all trait values serve as test cases.}
\label{fig:contradictory_example}
\end{figure}

To generate a solution's phenotype, we scan through all the genes in the genotype, identify the gene with the greatest value (ties are broken by picking the gene closest to the beginning of the genotype), and label it as the activation gene. 
The value of the activation gene is then directly copied into the trait residing in the same position within the phenotype and all other traits are set to zero (Figure \ref{fig:contradictory_example}).
This genotype-to-phenotype transformation enforces that a given individual cannot work towards optimizing multiple traits simultaneously.
As such, the complete set of optima consists of phenotypes with one trait set to the largest numerical value possible, while all other traits are set to zero.

This diagnostic favors selection schemes focused on finding Pareto-optimal solutions, as it generates a search space with multiple Pareto solutions.
In fact, \cite{hernandez2023suite} found that nondominated sorting excels on this diagnostic.
Selection schemes are challenged with maintaining a population consisting of solutions exploring multiple gradients while climbing these gradients to reach optima.
Lexicase can effectively explore without sacrificing its ability to exploit, but its ability to explore and maintain specialists across different test cases is influenced by the population size \citep{la2019probabilistic}.
When working with a fixed evaluation budget, the tradeoff between the population size and the number of generations will directly impact lexicase's ability to maintain multiple gradients within the population. 
Moreover, the number of generations may impact climbing progress a gradient. 

\subsection{Test case redundancy}
\label{sec:test-case-redundancy}


To our knowledge, all current work using the DOSSIER suite from \cite{hernandez2023suite} paired with lexicase selection incorporate exactly one test case for each trait within the phenotype.
This mapping from test case to trait creates an equal opportunity for each trait to appear first within the test case shuffling lexicase implements.
Here, we present the test case redundancy diagnostic, which generates duplicate test cases to measure how an imbalanced representation of traits affects lexicase's ability to maintain specialists.
The test case redundancy diagnostic is more of an extension, like the fitness-valley crossing component within \cite{hernandez2023suite}.
This diagnostic is particularly informative for lexicase selection, as prior work focuses on its ability to retain specialists for unique test cases \citep{la2019probabilistic}.

\begin{figure}[!h]
\centering
\includegraphics[width=0.8\textwidth]{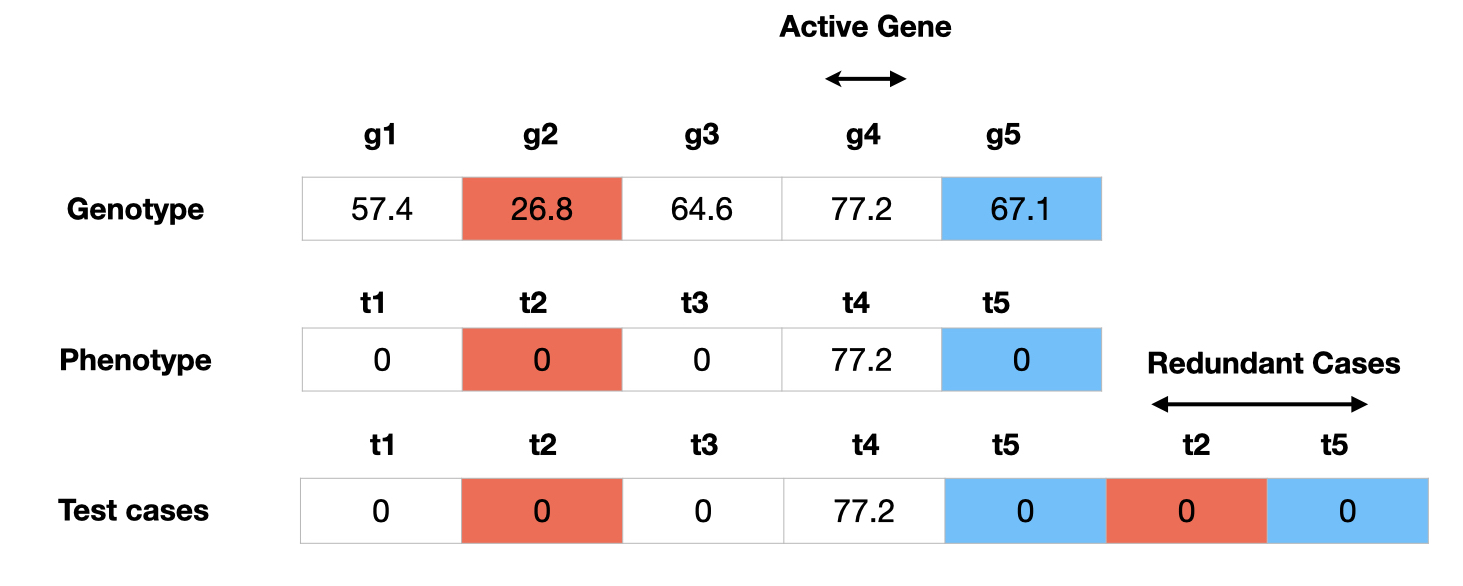}
\caption{Mapping from an example genotype to phenotype. The trait values corresponding to the colored genes are duplicated and placed at the end of the existing sequence of test cases.}
\label{fig:contradictory_redundant}
\end{figure}


This diagnostic measures how many specialists lexicase can maintain within the population when redundant test cases are integrated within the contradictory objectives diagnostic.
We extend the original set of test cases (the set of test cases in the standard contradictory objectives diagnostic) by appending redundant test cases. 
Each test case in the original set of test cases corresponds to a single trait within the phenotype, giving a total of $100$ test cases.
But here, multiple test cases can correspond to the same trait value giving us more than $100$ test cases.
Each additional test case is drawn from the original set of test cases with equal probability and replacement.
The new test case set is constructed at the start of an evolutionary run and lexicase selection will use it for each selection event. See Figure \ref{fig:contradictory_redundant} for an example.
In this work, we vary the number of additional test cases to $100$, $200$, and $400$.


\section{Methods}

We investigated how the combination of population size, test case redundancy, and number of generations under a consistent evaluation budget influences lexicase selection's ability to exploit gradients and maintain specialists.
From the DOSSIER suite in \cite{hernandez2023suite}, we used the exploitation rate diagnostic to measure exploitation and the contradictory objectives diagnostic to measure specialist maintenance.
Additionally, we extended the contradictory objectives diagnostic by incorporating redundancy within the number of test cases associated with a specific phenotypic trait.
We conducted five sets of experiments: one with the exploitation rate diagnostic, one with the contradictory objectives diagnostic, and three with the test case redundancy diagnostic.
For each experiment, we compared how varying population sizes ($50$, $100$, $500$, $1000$, and $5000$) impact performance with the given diagnostic.
In all, we had five unique diagnostics run with lexicase and a given population size, for a total of $25$ treatments.
We performed $50$ replicates within each treatment and restricted runs to a $1.5 \times 10^9$ evaluation budget.

\subsection{Evaluation budget}



\begin{table}[h]
  \caption{Number of generations lexicase selection with a specific population size and number of test cases can execute for a $1.5 \times 10^9$ evaluation budget. 
  Note that the standard version of the exploitation rate diagnostic and the contradictory objectives diagnostic are being considered when the redundancy of cases is set to $0$.
  For redundancy of test cases greater than $0$, the number of redundant cases is added within the contradictory objectives diagnostic.
  }
  \centering
  \begin{tabular}{|c|c|c|c|c|c|}
    \hline
    \multicolumn{2}{|c|}{} & \multicolumn{4}{c|}{Redundancy of cases} \\
    \cline{3-6}
    \multicolumn{2}{|c|}{} & 0 & 100 & 200 & 400 \rule{0pt}{2.5ex} \\
    \hline
    \multirow{6}{*}{\rotatebox[origin=c]{90}{Population size}} & 50 & 300,000 & 150,000 & 100,000 & 60,000 \rule{0pt}{2.8ex} \\
    \cline{2-6}
    & 100 & 150,000 & 75,000 & 50,000 & 30,000 \rule{0pt}{2.8ex} \\
    \cline{2-6}
    & 500 & 30,000 &  15,000 & 10,000 & 6,000 \rule{0pt}{2.8ex} \\
    \cline{2-6}
    & 1000 & 15,000 & 7,500 & 5,000 & 3,000 \rule{0pt}{2.8ex} \\
    \cline{2-6}
    & 5000 & 3,000 &1,500 & 1,000 & 600 \rule{0pt}{2.8ex} \\
    \hline
    \hline
    \multicolumn{2}{|c|}{Size of test case set} & 100 & 200 & 300 & 500 \rule{0pt}{2.6ex} \\
    \hline
  \end{tabular}
  \label{table:eval-gens}
\end{table}


In this work, we define the evaluation budget as the maximum number of times test cases are considered for parent selection throughout an evolutionary run.
The population size and the number of test cases used by lexicase selection determine the number of evaluations for a single generation.
For example, $100$ test cases and $100$ solutions will result in $10,000$ evaluations per generation.
For a fixed evaluation budget and fixed number of test cases, the population size determines the number of generations allowed during an evolutionary search.
We used an evaluation budget of $1.5 \times 10^9$ evaluations. 
As such, we adjusted the number of generations allowed for each population size used in this work.
Table \ref{table:eval-gens} demonstrates the number of generations for each lexicase configuration when evaluated on a specific diagnostic.

\subsection{Diagnostic experiments}


We set the dimensionality to $100$ for the diagnostics used in this work, resulting in $100$ genes and traits.
However, the number of total test cases used by lexicase selection varies by the diagnostic used.
For experiments with the exploitation rate diagnostic (Section \ref{sec:exploitation-rate-diagnostic}) and the contradictory objectives diagnostic (\ref{sec:contradictory-objectives-diagnostic}), the standard $100$ test cases are used to identify parent solutions (as in \cite{hernandez2023suite}).
For experiments with the test case redundancy diagnostic (Section \ref{sec:test-case-redundancy}), the total number of test cases varies by the number of redundant test cases added (100 sampled test cases generates 200 total test cases, 200 sampled test cases generates 300 total test cases, \textit{etc.}).
The bottom row of Table \ref{table:eval-gens} summarizes the total number of test cases for each redundancy count.


At the start of an evolutionary run, the initial population is comprised of solutions with genotypes consisting of random values between $0.0$ and $1.0$.
This process was used to start with a set of random solutions that reside near the lowest point in the search space.
Once the initial population is constructed, each solution is evaluated with the given diagnostic and assigned a phenotype.
Lexicase then uses the set of test cases to identify parents where the total number of identified parents is equal to the size of the initial population.
The number of solutions within the initial population depends on the population size configuration of the particular experiment.

Each identified parent reproduces asexually and there is a $0.07\%$ chance of a point mutation being applied to each gene within an offspring's genotype.
The magnitude of a point mutation is drawn from a Gaussian Distribution with the mean set to $0.0$ and the standard deviation set to $1.0$.
If a mutation results in a negative gene, we set the gene to its absolute value (\textit{e.g.}, a gene of $-0.7$ post-mutation would be adjusted to $0.7$).
If a mutation causes a gene to exceed the upper threshold ($100.0$), we rebound it by calculating the excess over the upper threshold and subtracting this excess from the upper threshold (\textit{e.g.}, a gene of $102.3$ post-mutation would be adjusted to $97.7$).
This rebounding process was used to evaluate lexicase's ability to refine solutions near optima.
The set of offspring make up the following population and the same cycle is repeated; the number of generations depends on the particular experiment configuration (Table \ref{table:eval-gens}).

\subsection{Data tracking}


The exploitation rate diagnostic measures how fast a selection scheme can hill-climb.
We collect performance data to determine how close a population has come to the optimum, and we use the total number of evaluations accumulated each generation to define time.
We calculated the average trait score (performance) for each solution's phenotype by summing all the traits and dividing the sum by the dimensionality ($100$); performance falls between $0.0$ and $100.0$.
We record the best performance found in the population each generation.
Additionally, we record the total number of evaluations accumulated when a satisfactory solution is found.
A solution is deemed satisfactory if the phenotype consists of all satisfactory traits, where a satisfactory trait is defined as a trait greater than or equal to $99.0$.
Recording when a satisfactory solution is found illustrates how efficiently lexicase can reach them.


All other diagnostics in this work focus on specialist maintenance under various conditions.
We collect activation genes from all solutions in the population and count the number of unique activation genes to calculate the activation gene coverage. 
This coverage value demonstrates lexicase's ability to maintain multiple specialists within the population, although they may not be satisfactory.
We collect all of the satisfactory traits found in the population and count the number of unique satisfactory traits to calculate satisfactory trait coverage.
This coverage value illustrates lexicase's ability to simultaneously push multiple specialists toward becoming satisfactory.
Activation gene coverage ranges between $1$ and $100$ and satisfactory trait coverage ranges between $0$ and $100$.
Note that redundant test cases do not alter these coverage ranges, as redundant test cases do not create new genes or traits.

\subsection{Statistical analysis}

We conducted a Kruskal-Wallis test to determine if significant differences occurred between the population sizes.
If comparisons resulted in significant differences with the Kruskal-Wallis test, we performed a post-hoc Wilcoxon rank-sum test to identify differences among population size pairings with a Bonferroni correction for multiple comparisons.
We used a significance level of $0.05$ for all statistical tests.

\subsection{Software availability}

Our supplemental material \citep{supplemental_material} is hosted on GitHub and contains all files related to software, data analysis, figure visualization, and documentation for this work.
We used Python to implement our experiments and R version 4 for statistical testing and data visualizations.
All data is available on the Open Science Framework at \href{https://osf.io/g5u9p/}{https://osf.io/g5u9p/}.
\section{Results and Discussion}


\subsection{Smaller population sizes facilitate faster exploitation}
\label{sec:results:exploitation}

\begin{figure}[!h]
\centering
\includegraphics[width=\textwidth]{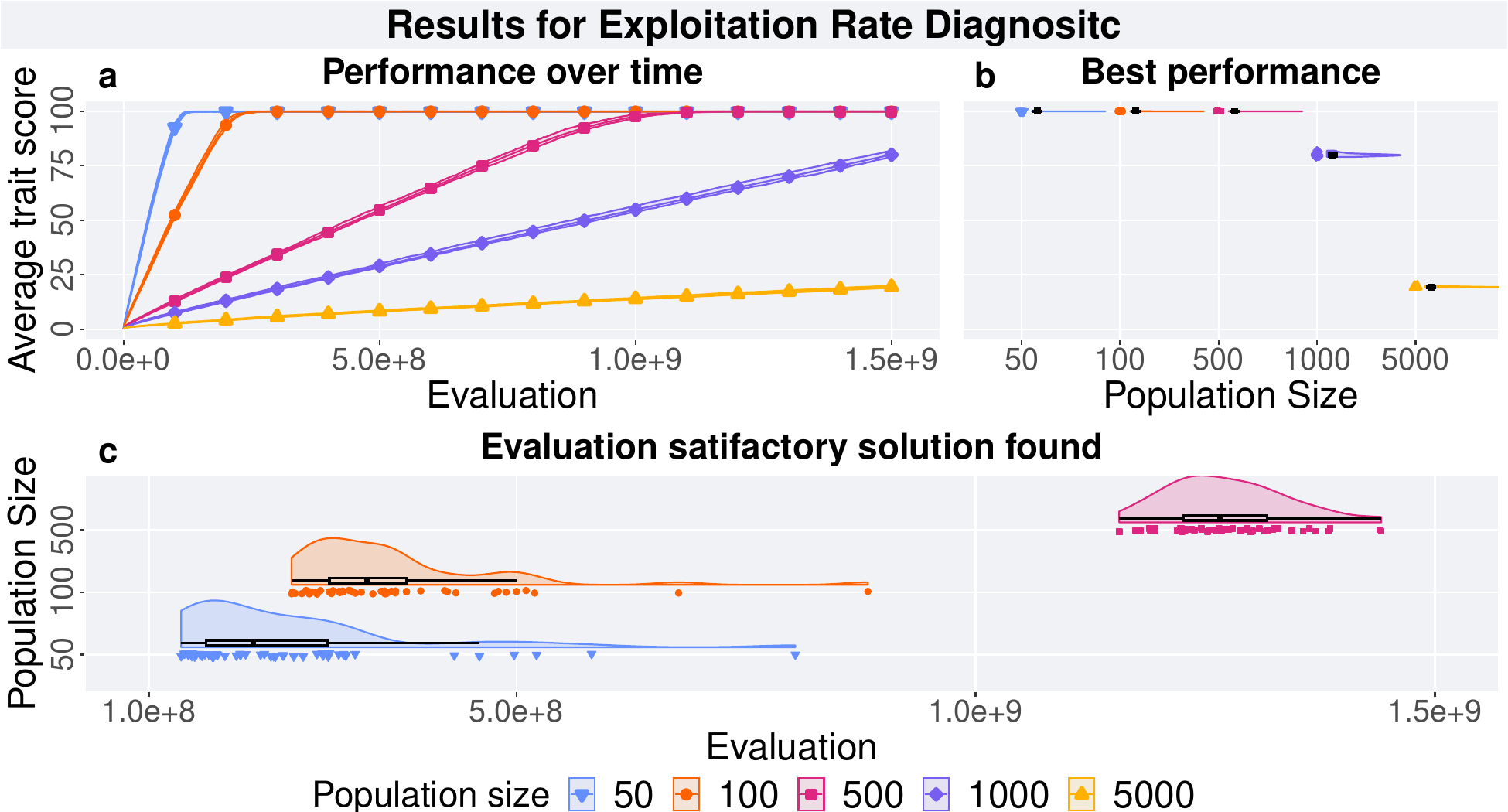}
\caption{
\textbf{Results for lexicase selection with varying population sizes on the exploitation rate diagnostic.}
We report (a) the best performance in the population over time, (b) the best performance evolved throughout the evolutionary run, and (c) the total accumulated evaluations when a satisfactory solution was first discovered.
For panel (a), we plot the average performance with surrounding boundaries from the best and worst performances across the 50 replicates for every $10^{8}$ evaluations.
}
\label{fig:res:exploitation}
\end{figure}

We used the exploitation rate diagnostic to measure how population size affects lexicase selection's ability to hill-climb given a fixed evaluation budget.
Figure \ref{fig:res:exploitation} illustrates the results for lexicase with varying population sizes on this diagnostic.
For the population sizes used in this work, only sizes $50$, $100$, and $500$ evolved satisfactory solutions within the given computational budget (Figure \ref{fig:res:exploitation}c).
We found that a population size of $500$ required more evaluations to find satisfactory solutions than both sizes of $100$ and $50$, and the population size of $50$ found satisfactory solutions faster than a size of $100$ (Wilcoxon rank-sum test: $p < 10^{-4}$).


Each configuration of lexicase improved performance over time (Figure \ref{fig:res:exploitation}a), albeit improving at different rates.
We can see that smaller population sizes exhibit faster improvement, with only sizes $50$, $100$, and $500$ reaching the optimum by the end of an evolutionary run.
The trajectory of improvement for population sizes of $1000$ and $5000$ indicates that they can reach optimum given a larger evaluation budget.
Figure \ref{fig:res:exploitation}b reports the best performance evolved throughout an evolutionary search.
Interestingly, we detected no difference when comparing these performances between sizes $50$ and $100$  (Wilcoxon rank-sum test: $p > 0.05$), and both of these sizes evolved better performances than all other sizes (Wilcoxon rank-sum test: $p < 10^{-3}$).
Among the remaining population sizes, a size of $5000$ resulted in the worst performances, while a size of $1000$ produced worse solutions than a size of 500 (Wilcoxon rank-sum test: $p < 10^{-4}$).

These results provide evidence that smaller population sizes facilitate more hill-climbing given a fixed evaluation budget.
Beneficial and harmful mutations have an equal probability of occurring, whereas solutions that accumulate beneficial mutations improve overall performance and the likelihood of becoming parents.
Conversely, solutions accumulating harmful mutations will likely not be identified as parents.
Given a fixed evaluation budget, smaller population sizes get more generations to evolve solutions, fewer selection events per generation, and more opportunities to accumulate beneficial mutations.
Larger population sizes, however, get fewer generations to evolve solutions and more selection events per generation, but at the expense of limited mutations accumulating over time.
Moreover, larger populations rely on encountering test case shuffles that identify overall top performers.

This diagnostic allows us to intuitively understand how the tradeoff between smaller and larger population sizes affects lexicase's ability to exploit.
To our knowledge, this is the first work looking directly at lexicase's ability to exploit and the selection pressure it exhibits toward top-performing solutions with varying population sizes.
Interestingly, \cite{ferguson2020characterizing} hypothesized that the downsampled variations of lexicase benefited from deeper evolutionary searches.
Here, we find additional evidence illustrating the advantages of using a smaller per-generation evaluation cost (\textit{i.e.} a smaller population size or a smaller set of test cases). 



\subsection{Larger population sizes yield more optimal specialists when no redundancy is added}
\label{sec:results:contradictory-100}

\begin{figure}[!h]
\centering
\includegraphics[width=\textwidth]{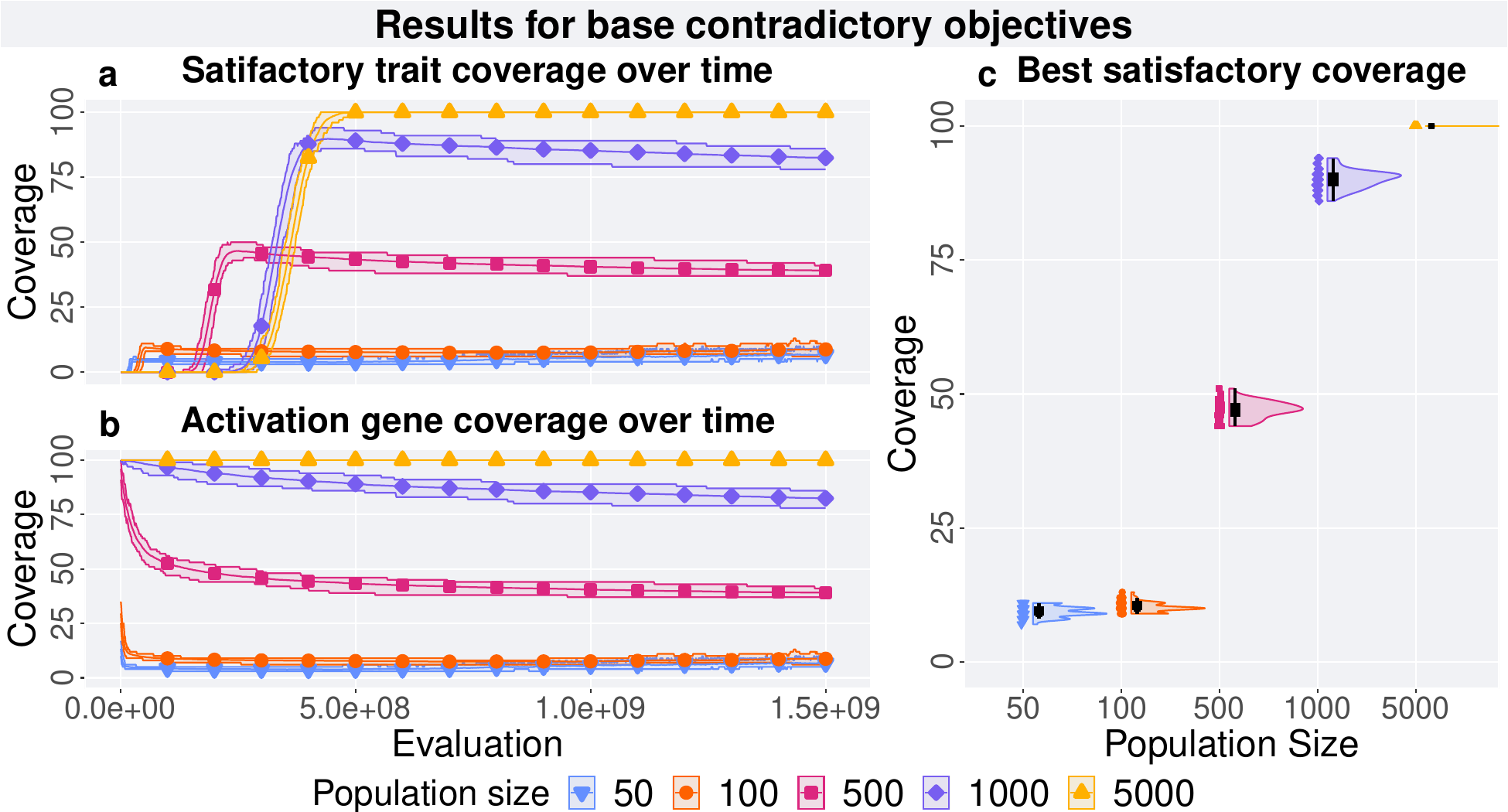}
\caption{
\textbf{Results for lexicase selection with varying population sizes on the contradictory objectives diagnostic.}
We report (a) the satisfactory trait coverage and (b) the activation gene coverage in the population over time, and (c) the best satisfactory trait coverage found throughout an evolutionary run.
For panels (a) and (b), we plot the average average with surrounding boundaries from the best and worst coverage across the 50 replicates for every $10^{8}$ evaluations.
}
\label{fig:res:base:contradictory}
\end{figure}

We used the contradictory objectives diagnostic to measure how population size affects lexicase selection's ability to maintain and optimize specialists when there is no redundancy of test cases.
Figure \ref{fig:res:base:contradictory} illustrates the results for lexicase on this diagnostic.
At the start of an evolutionary run, there is nearly perfect activation gene coverage for population sizes of $500$, $1000$, and $5000$, while the remaining sizes begin with low coverage (Figure \ref{fig:res:base:contradictory}b).
High activation coverage is expected with large sizes as the initial population consists of randomly generated genotypes with genes ranging between $0.0$ and $1.0$.
Interestingly, the population size of $5000$ maintains perfect coverage throughout the entire evolutionary run for all replicates.
The remaining sizes, however, see a decline in activation gene coverage and converge to different coverage values at the end of an evolutionary run.


Activation gene coverage illustrates how many specialists lexicase has maintained in the population, while satisfactory trait coverage demonstrates how many of those specialists reach satisfactory levels.
Figure \ref{fig:res:base:contradictory}a demonstrates the growth in satisfactory trait coverage over time.
Interestingly, the smaller population sizes reach satisfactory traits faster than larger sizes.
This result is consistent with the results for the exploitation rate diagnostic (Section \ref{sec:results:exploitation}), where smaller population sizes hill-climb quicker.
While the smaller sizes reached satisfactory traits quickly, they did not reach high coverage like larger sizes.
Panel (c) in Figure \ref{fig:res:base:contradictory} reports the best satisfactory trait coverage throughout an evolutionary run.
Interestingly, the population size of $5000$ evolves populations with perfect satisfactory trait coverage ($100$) for each replicate. 
All other population sizes reached lower coverage, where larger sizes reached better satisfactory trait coverage than smaller sizes  (Wilcoxon rank-sum test: $p < 10^{-2}$).


Overall, our results are consistent with previous work illustrating that lexicase can maintain more specialists with larger populations \citep{la2019probabilistic,hernandez2022exploration}.
Interestingly, all possible specialists were optimized to satisfactory levels with a population size of $5000$.
These results provide additional evidence that illustrates lexicase's ability to maintain specialists is determined by the population size. 

\subsection{Intermediate levels of redundancy restrict specialist optimality with large population sizes}

\begin{figure}[!h]
\centering
\includegraphics[width=\textwidth]{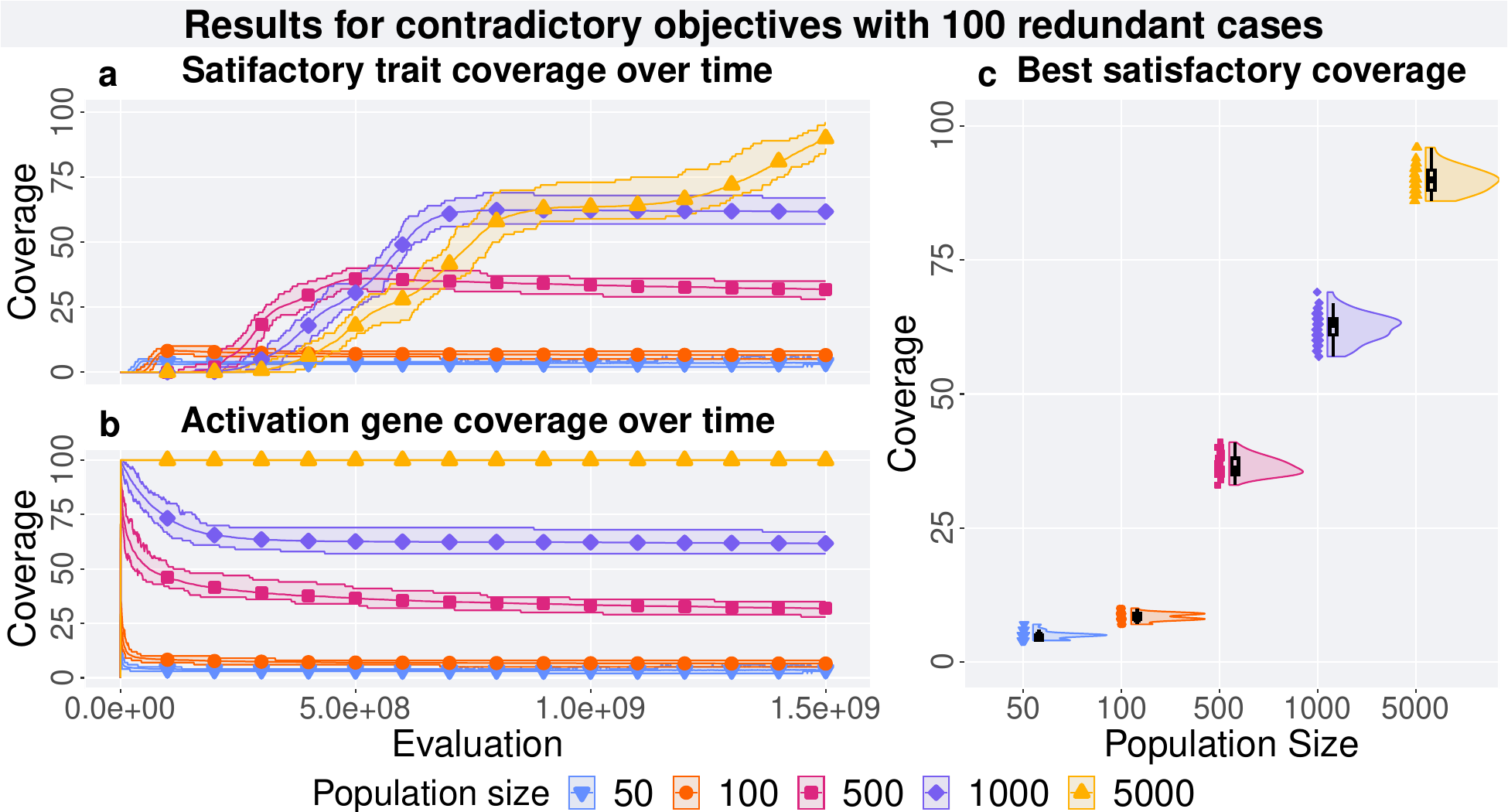}
\caption{
\textbf{Results for contradictory objectives with 100 redundant test cases} 
We report (a) the satisfactory trait coverage and (b) the activation gene coverage in the population over time, and (c) the best satisfactory trait coverage found throughout an evolutionary run.
For panels (a) and (b), we plot the average average with surrounding boundaries from the best and worst coverage across the 50 replicates for every $10^{8}$ evaluations.
}
\label{fig:res:100:contradictory}
\end{figure}
\begin{figure}[!h]
\centering
\includegraphics[width=\textwidth]{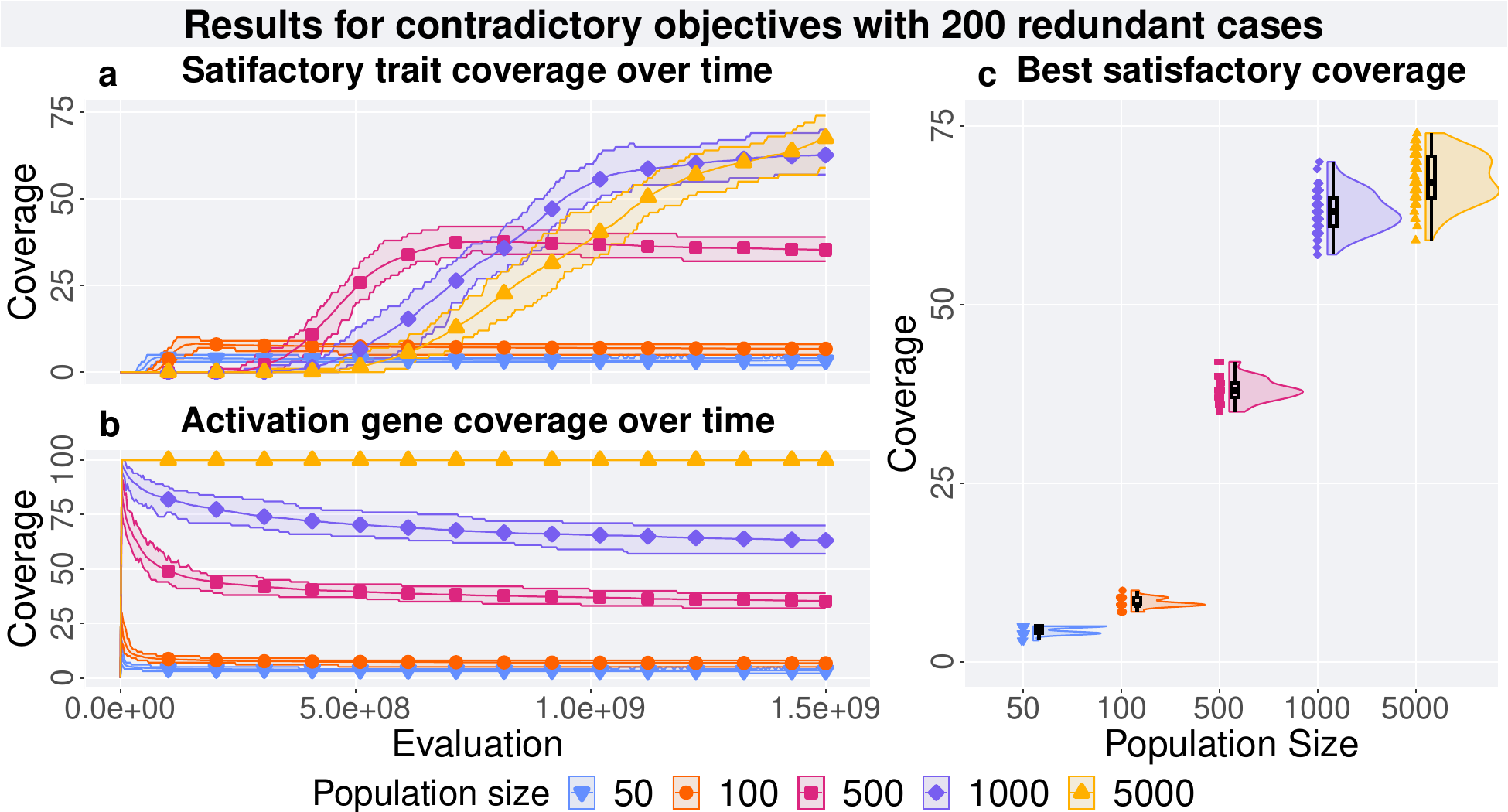}
\caption{
\textbf{Results for contradictory objectives with 200 redundant test cases} 
We report (a) the satisfactory trait coverage and (b) the activation gene coverage in the population over time, and (c) the best satisfactory trait coverage found throughout an evolutionary run.
For panels (a) and (b), we plot the average average with surrounding boundaries from the best and worst coverage across the 50 replicates for every $10^{8}$ evaluations.
}
\label{fig:res:200:contradictory}
\end{figure}

Incorporating redundancy through duplicate test cases alters the selection dynamic of lexicase selection.
In this section, we discuss the results of adding 100 and 200 redundant test cases to the set of test cases used by the lexicase. 
The results for 100 redundant cases are shown in Figure \ref{fig:res:100:contradictory}, while the results for 200 redundant cases are displayed in Figure \ref{fig:res:200:contradictory}.


Interestingly, adding either 100 or 200 redundant test cases produced similar results.
In both cases, larger population sizes start with high activation gene coverage.
However, coverage falls for most sizes within a few generations. 
The population size of 5000 maintained high coverage throughout the entire evolutionary run.
In contrast, sizes 1000 and 500 maintain moderate coverage, while sizes 50 and 100 maintain low coverage. 
Regarding satisfactory trait coverage, the population size of 5000 maintained the most coverage and continuously improved coverage over time. 
For sizes 1000 and 500, there is an increase in satisfactory trait coverage over time, but the coverage eventually plateaus.
The population sizes 50 and 100 have poor coverage throughout the evolutionary run. 
At the end of the respective runs, the population size of 5000 produced the populations with the best satisfactory coverage, followed by larger sizes producing more coverage than smaller sizes (Wilcoxon rank-sum test for all pairs; $p<10^{-4}$) 


Looking at Figure \ref{fig:res:100:contradictory}c (redundancy of 100) and Figure \ref{fig:res:200:contradictory}c (redundancy of 200), one clear trend emerges.
The best satisfactory coverage decreases with an increase in redundancy. 
One explanation for this behavior is that with redundancy in test cases, the probability of non-duplicate test cases appearing at the beginning of the shuffle in lexicase decreases.
As a result, larger population sizes are needed to increase the chances of all test cases appearing first in the shuffle.
Let us take an example to illustrate this point. Consider this sequence of test cases: $a,b,c,d$. 
Without redundancy, every test case has an equal probability of coming first in the shuffle. 
Consider the same sequence with redundancy added: $a,a,b,b,c,d$. 
Clearly, $a$ and $b$ are more probable to come first in the shuffle than $c$ and $d$. Therefore, lexicase selection is more likely to maintain specialists for $a,b$. 

\subsection{Intermediate population sizes produce a greater number of optimal specialists with numerous redundant test cases}

\begin{figure}[!h]
\centering
\includegraphics[width=\textwidth]{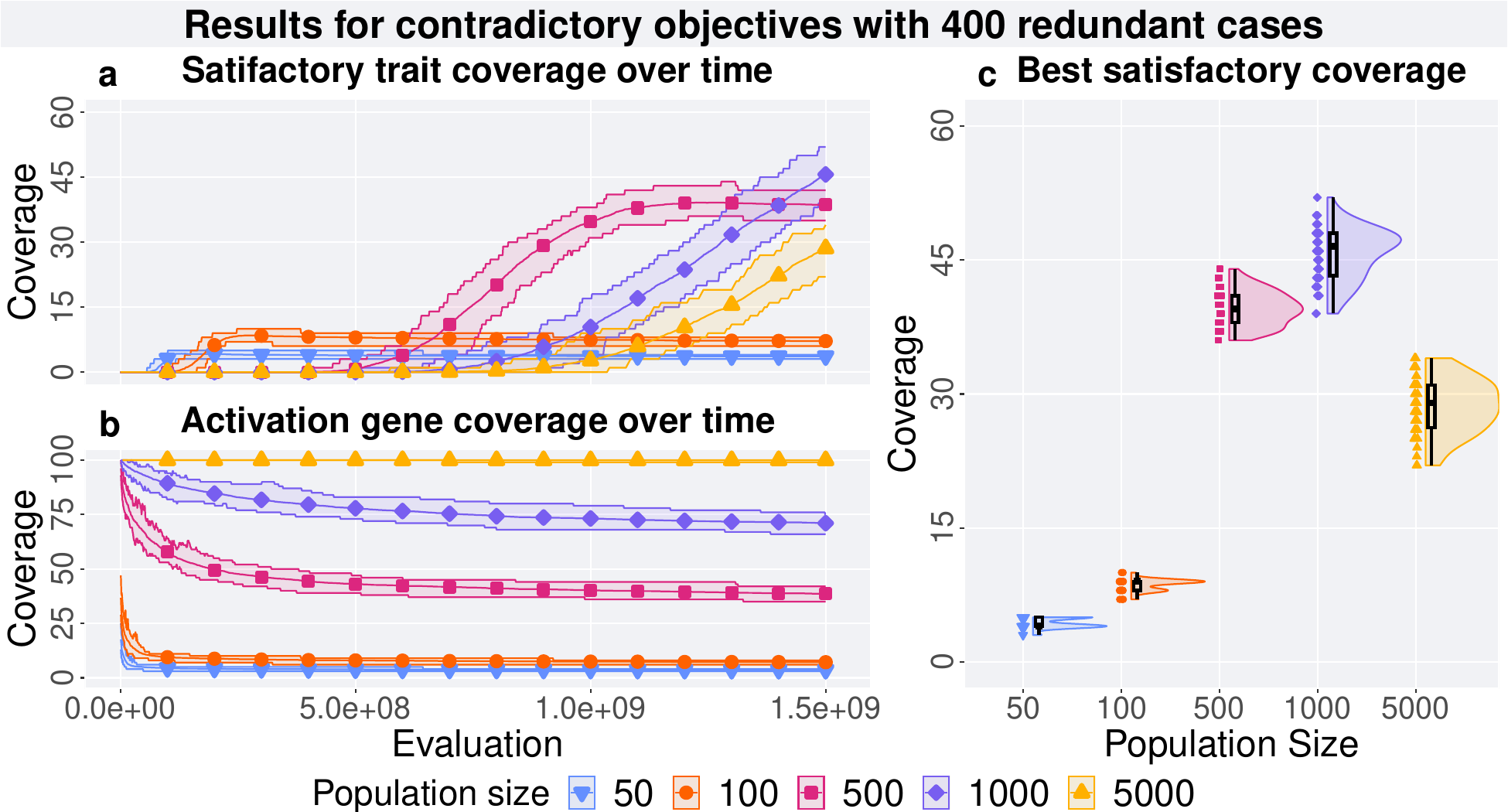}
\caption{
\textbf{Results for lexicase selection with varying population sizes on the contradictory objectives diagnostic with $400$ redundant test cases.}
We report (a) the satisfactory trait coverage and (b) the activation gene coverage in the population over time, and (c) the best satisfactory trait coverage found throughout an evolutionary run.
For panels (a) and (b), we plot the average average with surrounding boundaries from the best and worst coverage across the 50 replicates for every $10^{8}$ evaluations.
}
\label{fig:res:400:contradictory}
\end{figure}

Figure \ref{fig:res:400:contradictory} illustrates how lexicase selection performed on the test case redundnacy diagnostic with our most extreme level of redundancy -- $400$ redundant cases.
Interestingly, the population size of $5000$ excelled with all smaller levels of redundancy, reaching higher satisfactory trait coverage than all other sizes (Wilcoxon rank-sum test: $p < 10^{-4}$).
With the $400$ redundant test cases, however, the population size of $5000$ does not achieve the greatest satisfactory trait coverage. 

The activation gene coverage maintained by the different population sizes follows a similar trend to the previous results (Figure \ref{fig:res:400:contradictory}b).
The population size of $5000$ maintained nearly perfect activation gene coverage throughout the entire run for all replicates.
The remaining population sizes maintained less coverage, where larger sizes maintained more coverage than smaller sizes.
While the size of $5000$ managed to maintain nearly perfect activation gene coverage, it was unable to reach high satisfactory trait coverage like in the previous results.
In fact, the population size of $5000$ reached lower satisfactory trait coverage than both sizes $1000$ and $500$, and the size of $1000$ reached greater coverage than the size of $500$ (Wilcoxon rank-sum test: $p < 10^{-4}$).
The remaining population sizes maintained less coverage, with size $100$ generating more coverage than size $50$ (Wilcoxon rank-sum test: $p < 10^{-4}$).


Figure \ref{fig:res:400:contradictory}a illustrates the growth in satisfactory trait coverage over time for the different population sizes.
Interestingly, the population sizes of $5000$ and $1000$ appear to improve satisfactory trait coverage throughout the evolutionary run, whereas all other sizes plateau.
The high activation gene coverage maintained by the size $5000$ (\ref{fig:res:400:contradictory}b) and the growth trajectory in satisfactory trait coverage (\ref{fig:res:400:contradictory}a) suggest that it might eventually reach perfect satisfactory trait coverage, surpassing all other sizes.
The per-generation evaluation cost is now a crucial aspect to consider given that $500$ test cases must now be considered.
For larger population sizes, the per-generation evaluation cost is $250,000$ for a size of $500$, $500,000$ for a size of $1000$, and $2,500,000$ for a size of $5000$.
These results present a clear tradeoff between using different population sizes for maintenance and optimizing specialists with an evaluation budget.
\section{Conclusion}

In this work, we used the DOSSIER suite in \cite{hernandez2023suite} to examine how lexicase selection's capacity for exploitation and specialist maintenance is affected by different population sizes given a consistent evaluation budget.
We used the exploitation rate diagnostic to capture lexicase's ability to exploit and the contradictory objectives diagnostic to capture lexicase's ability to maintain specialists.
Additionally, we extended the contradictory objectives diagnostic by incorporating redundant test cases into the pool of test cases lexicase shuffles to consider a new dynamic regarding specialist maintenance.
Ultimately, these diagnostics enable us to dissect lexicase selection under various conditions and provide an intuitive understanding of why each condition yields a particular effect. 
For future work, it would be interesting to compare how more complex variants of lexicase selection differ from one another (\textit{e.g.}, downsampled lexicase). 


Lexicase selection is typically presented as being parameter-less when compared against other selection schemes, as all the parameters used by an evolutionary algorithm are predefined (\textit{e.g.} population size, variation operators, mutation rate, \textit{etc.}).
However, when comparing lexicase to itself with a fixed computational budget and problem, it is clear that the population size is a varying parameter that can affect its performance.
All practitioners implicitly use an evaluation budget (the number of times test cases are considered during parent selection) when using lexicase selection, whether they are aware of it or not.
As such, the question becomes what is the ideal parameter setting for lexicase selection given a problem being solved and an evaluation budget?


Our results illustrate that there is no perfect population size for all problems and the characteristics of a given problem must be considered for the best configuration of the evolutionary algorithm paired with lexicase selection.
The results of the exploitation rate diagnostic present evidence that smaller population sizes should be preferred for problems that require exploitation to reach optima.
To our knowledge, this is the first work analyzing how population size affects lexicase selection's ability to exploit and steer a population toward optima.
Conversely, the results of the contradictory objectives diagnostic present evidence that larger population sizes should be preferred for problems that require specialists to be maintained to reach optima.
This result is consistent with previous work illustrating that lexicase's ability to maintain specialists is determined by the population size being used. 


We considered a new aspect regarding lexicase selection's ability to maintain specialists.
Previous works often assume that a unique test case must come first in the shuffle to allow a specialist to be selected.
Here, however, we examine the scenario where duplicate test cases exist that allow a specialist to be selected by allowing duplicate test cases within the contradictory objectives diagnostic.
The addition of duplicate test cases creates a new dynamic to be considered as the probability of a specialist being selected now varies and the per-generation evaluation cost is increased.
We found that larger population sizes can maintain more specialists, with the largest population size ($5000$) used in this study performing the best.
However, we found the largest population size was outperformed by smaller sizes ($1000$ and $500$) when $400$ redundant test cases were used.
These results highlight the importance of considering on how many test cases to use for a given problem and the information each test case captures to guide the evolutionary search.
Clearly, having correlated test cases may lead to an excessive number of test cases to be considered, potentially wasting evaluations when working within a constrained budget.




\bibliographystyle{apalike}
\bibliography{references,software}

\end{document}